\documentclass[final,3p]{elsarticle}

\usepackage{graphicx}
\usepackage{hyperref}
\usepackage[utf8]{inputenc}

\usepackage[T1]{fontenc}    

\usepackage{url}            
\usepackage{booktabs}       
\usepackage{amsfonts}       
\usepackage{nicefrac}       
\usepackage{microtype}      
\usepackage{xcolor}         

\usepackage[ruled,vlined]{algorithm2e}
\usepackage{amsmath}
\DeclareMathOperator*{\argmin}{arg\,min}
\usepackage{amsthm}
\usepackage{amssymb}

\usepackage{amsfonts}
\newcommand{\round}[1]{\ensuremath{\lfloor#1\rceil}}

\begin{document}

\title{The Complexity Dynamics of Grokking}

\author[1]{Branton DeMoss\corref{cor1}}
\ead{bdemoss@robots.ox.ac.uk}

\author[1]{Silvia Sapora}

\author[1]{Jakob Foerster}

\author[1]{Nick Hawes}

\author[1]{Ingmar Posner}

\cortext[cor1]{Corresponding author}
\address[1]{Department of Engineering Science, University of Oxford, Oxford, United Kingdom}

\begin{abstract}
We demonstrate the existence of a complexity phase transition in neural networks by studying the grokking phenomenon, where networks suddenly transition from memorization to generalization long after overfitting their training data. To characterize this phase transition, we introduce a theoretical framework for measuring complexity based on rate-distortion theory and Kolmogorov complexity, which can be understood as principled lossy compression for networks. We find that properly regularized networks exhibit a sharp phase transition: complexity rises during memorization, then falls as the network discovers a simpler underlying pattern that generalizes. In contrast, unregularized networks remain trapped in a high-complexity memorization phase. We establish an explicit connection between our complexity measure and generalization bounds, providing a theoretical foundation for the link between lossy compression and generalization. Our framework achieves compression ratios 30-40$\times$ better than naïve approaches, enabling precise tracking of complexity dynamics. Finally, we introduce a regularization method based on spectral entropy that encourages networks toward low-complexity representations by penalizing their intrinsic dimension.
\end{abstract}


\maketitle

\section{Introduction}

When can we expect learned models to generalize well? The study of phase transitions in learning systems provides a window into fundamental questions about the nature of generalization and emergence of structure. In this work, we investigate the recently observed “grokking” phenomenon \citep{grokking}, where neural networks exhibit a sharp transition from memorization to generalization long after overfitting their training data. The extended delay between these two phases of learning provides an ideal setting to study the relationship between complexity and generalization in learning systems.

Simplicity has long been recognized as a common feature of good models: Occam's Razor says that amongst models which explain some data equally well, the simplest one is expected to generalize best. Drawing inspiration from statistical physics and algorithmic information theory, we introduce a framework for measuring the intrinsic complexity of neural networks based on lossy compression and Kolmogorov complexity. Our approach parallels work by \cite{aaronson2014} which studies the complexity dynamics of physical systems. Their key insight is to measure complexity after appropriate \textit{coarse-graining} of the system's state. We formalize their insight using algorithmic rate--distortion theory, resulting in a principled lossy compression scheme for neural networks which bounds their complexity. 

\begin{figure}
    \centering
    \includegraphics[width=\linewidth]{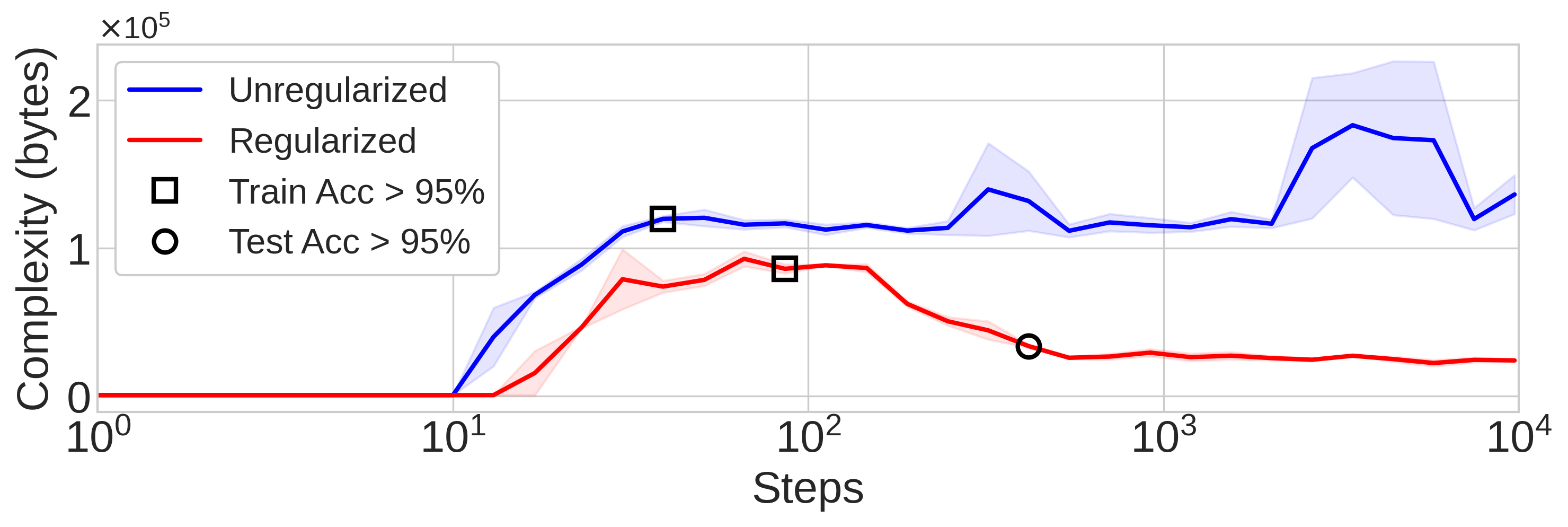}
    \caption{Complexity dynamics for unregularized (blue) and regularized (red) neural networks in grokking experiments on modular multiplication. Markers denote when memorization ($\square$) and generalization ($\bigcirc$) occur. In the regularized network, generalization occurs as complexity falls after its peak at memorization, while the unregularized network never generalizes, and its complexity remains large. Mean over 3 seeds, shading std. error.}
    \label{fig1}
\end{figure}

This framework allows us to track the complexity dynamics of neural networks throughout training. We find that properly regularized networks that successfully generalize exhibit a distinct phase transition: complexity first rises as the network memorizes training data, then falls sharply as the network discovers a simpler underlying pattern that generalizes. In contrast, unregularized networks that fail to generalize remain trapped in a high-complexity memorization phase indefinitely. We establish an explicit link between our complexity measure and statistical generalization bounds, and show how these correspond to the Minimum Description Length Principle.

The Minimum Description Length (MDL) Principle \citep{Rissanen1978} formalizes Occam's Razor in the language of information theory. It says that the best model is the one which minimizes the total message length of the data. Shannon's theorem \citep{Shannon1948} says that the expected optimal coding rate for samples drawn from a distribution is equal to the entropy of the distribution. Hence, the entropy $H$ of some data $D$ under a model $M$, $H(D|M)$ tells us the expected message length after encoding the data with the model. However, we must also consider the information cost of the model itself, which is the model complexity $C(M)$. MDL says we should take the model which minimizes the \textit{sum} of these terms:

\begin{equation}\label{eq:mdl}
    M \leftarrow \argmin_M H(D \mid M) + C(M)
\end{equation}

Intuitively, it is worth making our model one bit more complex if it lets us reduce the apparent entropy of the data by more than one bit, since that reduces the overall message length. Note that if the model $M$ fully explains some subset $D' \subseteq D$ then the prediction of that data becomes deterministic and $H(D'|M) = 0$. Consider a trivial model which simply memorizes the data in a lookup table. Whilst this reduces the entropy of the data to zero, the model complexity is exactly equal to the original entropy of the data. Therefore the total description length has not changed, and no compression has been achieved by the model.

The insight that simpler models are expected to generalize better is made precise through statistical generalization bounds of the form:
\begin{equation}
\label{generalizationcomplexitybound}
    \text{test error} \leq \text{train error} + \text{model complexity}
\end{equation}
which are ubiquitous across a number of fields of statistical machine learning \citep{Vapnik1991, Rademacher}. When the error terms are identified with entropy, as is typically the case, these bounds reveal that models which minimize MDL are precisely those expected to generalize best. This provides the link between compression and generalization: the model which best compresses the data is the one expected to generalize best.

To apply the MDL principle, we require a measure of model complexity $C(M)$. Traditional complexity measures in machine learning such as Rademacher complexity \citep{Rademacher} or VC dimension \citep{VCDimension} are defined relative to specific hypothesis classes, but we seek a universal measure that applies to any computational model. The Kolmogorov complexity $K$ of a string $s$ is equal to the length of the shortest program $p$ which prints $s$ when executed:
\begin{equation}
\label{kcomplexitydefinition}
    K(s) = \min_p \{\texttt{len}(p) : \texttt{exec}(p) = s \}
\end{equation}
To build intuition, consider two strings of length $N$: The string `$111...1$' can be generated by a short program \texttt{print `1' N times}, so $K(s) \approx \log N$. In contrast, a random string `$10110...$' has no pattern to exploit; its shortest description is itself: \texttt{print `10110...'}, giving $K(s) \approx N$. Algorithmically random strings are precisely those whose Kolmogorov complexity equals their length—they are incompressible because they contain no exploitable patterns. 
For neural networks, this means a network implementing a simple pattern should have low complexity despite having many parameters, while a network memorizing random labels must have high complexity. Unlike capacity measures that provide worst-case upper bounds, Kolmogorov complexity captures the actual information content of the learned function.

\begin{figure}
    \centering
    \includegraphics[width=0.75\linewidth]{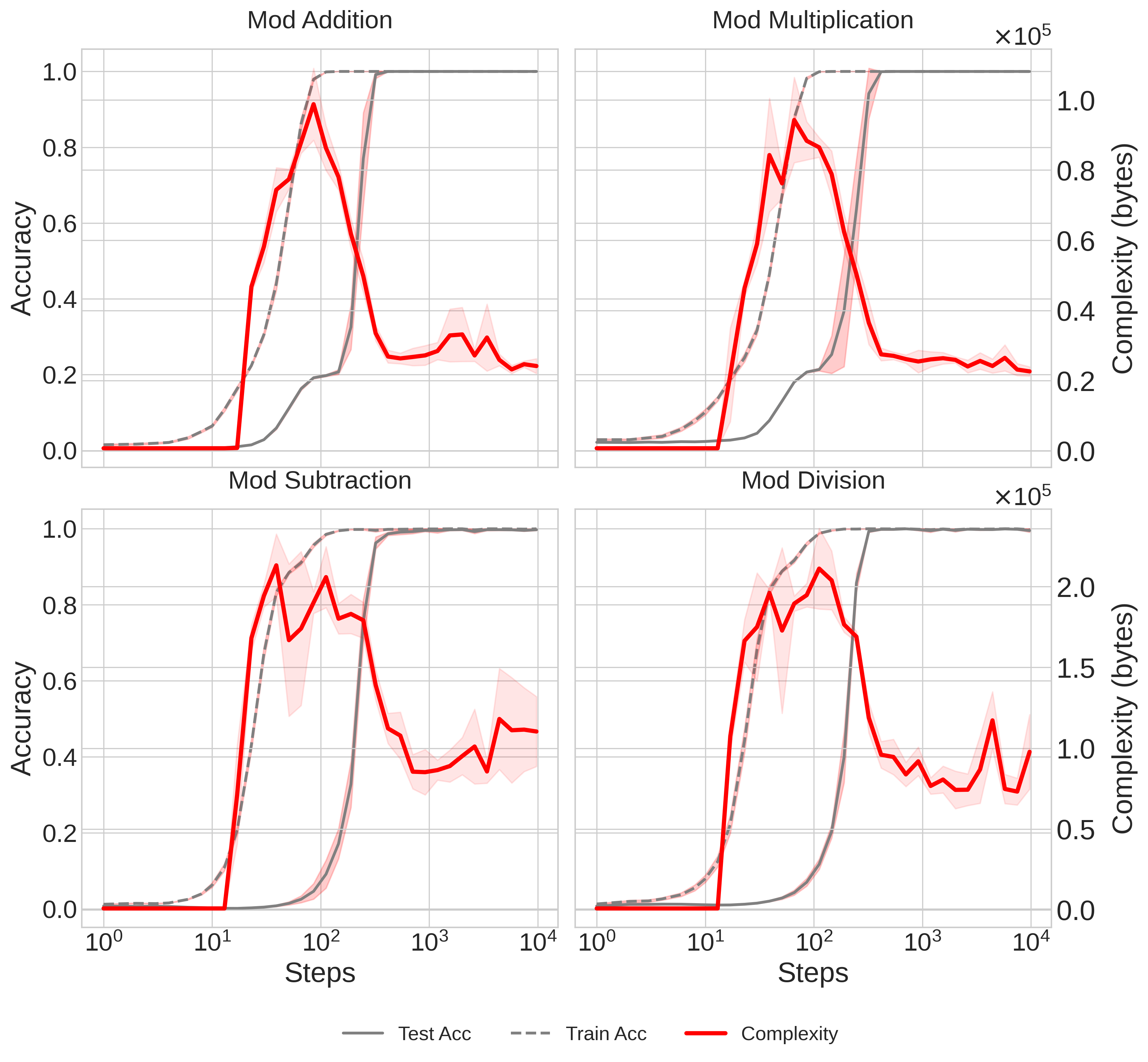}
    \caption{Complexity and accuracy vs steps for models regularized using our method. As models memorize, complexity increases. Complexity falls as generalization occurs. We use Algorithm \ref{alg:complexity2} to coarse-grain models with $\epsilon=1$. Std. error shaded with six seeds.}
    \label{complexaccuracy}
\end{figure}

To implement these ideas, we develop a principled approach to measuring neural network complexity based on lossy compression. While the Kolmogorov complexity is uncomputable, compression provides an upper bound: notice that compressed data plus its decompressor form a program that generates the original data. The compressed file-size of the network provides an upper bound on its complexity, with better compression resulting in tighter upper bounds. To compute tight compression bounds, we develop a coarse-graining procedure inspired by rate-distortion theory that removes noise while preserving functional behavior, enabling compression ratios 30-40× better than naïve approaches. This sensitivity is crucial for revealing the underlying complexity dynamics during training, shown in Fig~\ref{fig1}. Our main contributions are:
\begin{itemize}
    \item A theoretical framework connecting neural network complexity to Kolmogorov complexity through lossy compression, with explicit links to generalization bounds.
    \item Empirical demonstration of a complexity phase transition during grokking, characterized by a rise and fall pattern that distinguishes generalizing from memorizing solutions.
    \item A regularization method based on spectral entropy that encourages networks toward low-complexity representations by penalizing a differentiable measure of their effective dimension.
\end{itemize}

\begin{figure}
    \centering
    \includegraphics[width=0.75\linewidth]{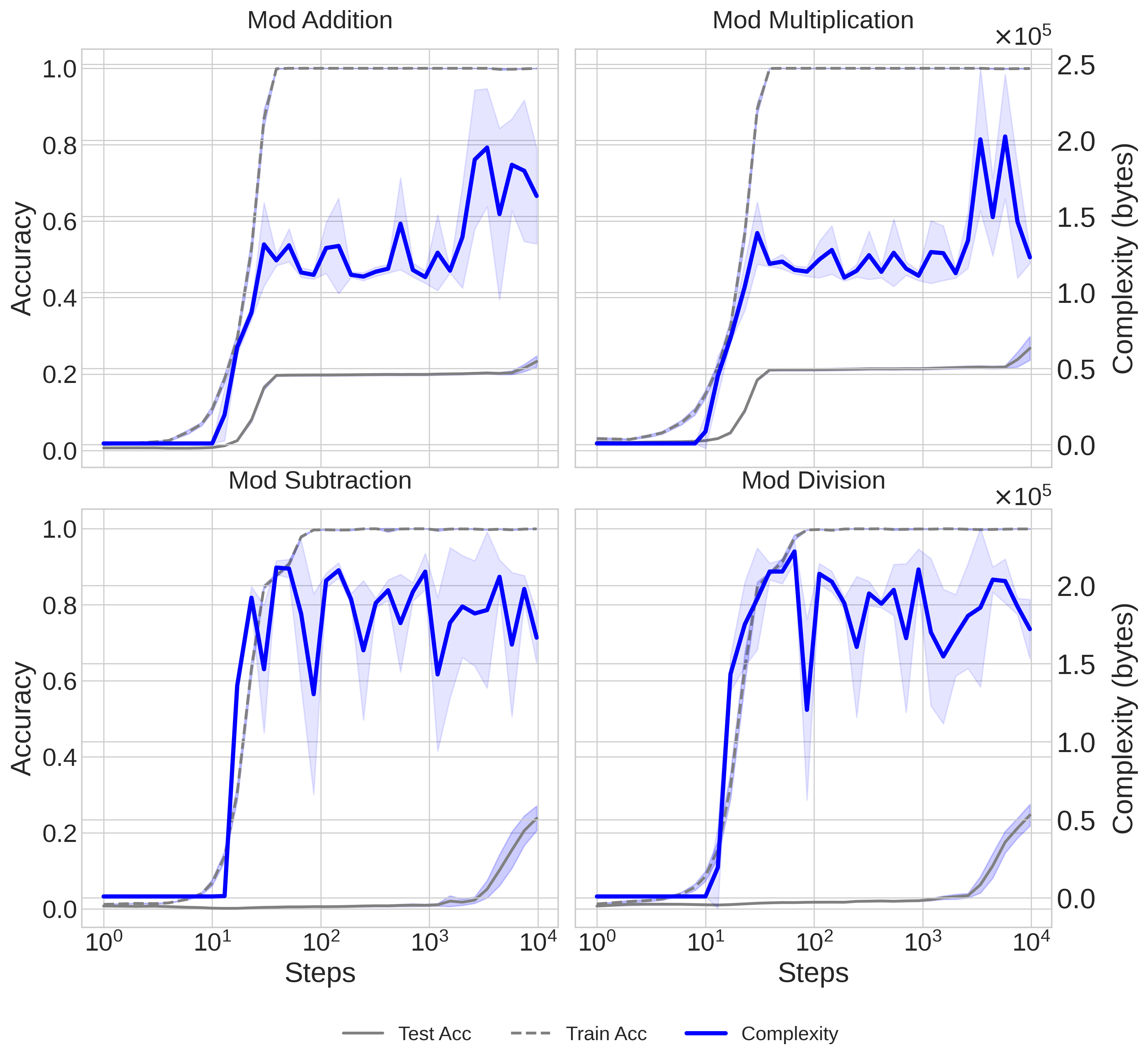}
    \caption{Complexity and accuracy vs steps for unregularized models. As models memorize, complexity increases. Without regularization, the models fail to find low complexity solutions, and never generalize. We use Algorithm \ref{alg:complexity2} to coarse-grain models with $\epsilon=1$.}
    \label{unregularizedcomplexity}
\end{figure}

\section{Related Work}
\textbf{Grokking}\hspace{0.5cm} The grokking phenomenon \citep{grokking} occurs when networks suddenly transition from memorizing to generalizing solutions long after over-fitting the training data. \cite{liu2022} characterize conditions under which grokking occurs, and categorize learning into four distinct regimes, \textit{comprehension}, \textit{grokking}, \textit{memorization}, and \textit{confusion} by examining phase diagrams of learning performance across different hyperparameters. \cite{omnigrok} demonstrate grokking on non-algorithmic datasets, and propose a link between a network's parameter norm and grokking. This link is also explored by \cite{varma2023}, who explain grokking in terms of the ``efficiency'' of a network, as measured by the ratio of the logit to parameter norms. \cite{Nanda2023} propose task-specific loss-based ``progress measures'' to detect grokking, but note that we ``lack a general notion of criticality that would allow us to predict when the phase transition'' occurs. In this work, we suggest that complexity is the key dynamical metric to understand grokking (and generalization more broadly), and discuss why proxy measures such as the $L^2$ norm are insufficient.

\textbf{Generalization}\hspace{0.5cm} The connection between model capacity and complexity is sometimes misunderstood amongst machine learning practitioners: the contemporary view in deep learning is that ``larger models are better'' \citep{deepdoubledescent}, with empirically observed scaling laws across a range of modalities and architectures \citep{kaplan2020, henighan2020} which show monotonic improvement of performance with increasing model scale. This view contrasts with the classical bias-variance tradeoff, which says that ``once we pass a certain threshold, larger models are worse'' \citep{deepdoubledescent}. The latter view is only true assuming model capacity is equal to model complexity, and would follow from bounds of the form of Equation \ref{generalizationcomplexitybound}. However, this appears \textit{not} to be the case. \cite{Lotfi2024} recently established the first non-vacuous generalization bounds for large language models by producing highly compressible models trained in a non-linear low-rank subspace, and bounding their generalization performance under a smoothed entropy loss by the Kolmogorov complexity. While they focus on explicit generalization \textit{bounds}, we focus on the complexity dynamics of the transition from memorization to generalization to explain grokking. \cite{Goldblum2023} argue for generalization to be understood using Kolmogorov complexity, and apply the Solomonoff prior, $P(h) = 2^{-K(h)}$ for hypotheses $h$ to a finite hypothesis bound \citep{finitehypothesis} to produce a generic generalization bound based on the Kolmogorov complexity.

\textbf{Coarse-graining}\hspace{0.5cm} \cite{aaronson2014} study the complexity dynamics of a cellular automaton which simulates the mixing of two liquids. They observe the rise and fall of complexity as entropy increases, and conjecture that this phenomenon is generic. To approximate the complexity of their automaton, they propose an ``apparent complexity'' measure which is the compressed file size of a \textit{coarse-grained} description of the state. They informally assert that coarse-graining removes random information from the state, leaving only ``interesting'' information behind. We formalize their intuition using rate--distortion theory, and use the resultant complexity measure to explain grokking.

\section{Complexity and Generalization}\label{section3}
Understanding generalization is of central importance in machine learning. 
To what extent can we expect our models to generalize, and can we predict their performance ahead of time? 
Most generalization bounds are a function of model complexity, so understanding complexity is key to understanding generalization. 
\cite{Lotfi2024} provide one such bound, which quantifies the expected risk $R(h)$ for a given hypothesis $h$, with probability $1-\delta$:
\begin{equation}\label{genbound}
    R(h) \leq \hat{R}(h) + \sqrt{\frac{K(h)+2\log K(h) + \log (1/\delta)}{2n}}
\end{equation}
Where $K(\cdot)$ is the Kolmogorov complexity, $\hat{R}(h)$ is the empirical risk, and $n$ the number of samples. 
As mentioned in the introduction, the Kolmogorov complexity is not computable, but can be upper-bounded by compression. 
Therefore, we can bound our generalization error by compressing models, with better bounds following from tighter compression. 
It is important to keep in mind that some approaches we might take to compress models, such as reducing the precision of their parameters, can also result in loss of performance. 
Hence, while we might decrease the complexity term by compressing models, this may result in the empirical risk $\hat{R}(h)$ increasing. 
To produce strong generalization guarantees, we must jointly minimize the model complexity and the empirical risk.

In fact, this can also be understood as a restatement of the MDL principle. 
If the empirical risk is understood as the entropy of the training data under the model, as is typically the case, then the joint minimization of the empirical risk and model complexity is exactly the MDL principle. 
I.e. the model which generalizes best is the one which compresses the data best, and is itself most compressible.

Unfortunately, na\"ively compressing a model's parameters produces poor complexity bounds, since the network contains random information left over from initialization, and from the stochastic training procedure. 
If we could remove the random information from a network, compressing the resulting parameters would give much tighter bounds on the network complexity. 
However, it is undecidable whether information is random \citep{LiVitanyi}, so there is no generic method to remove it.

\cite{aaronson2014} remove random information by coarse-graining their automaton's state. They coarse-grain by smoothing and quantizing the state. 
The coarse-grained state is then compressed with off-the-shelf programs such as \texttt{bzip2}, with the compressed file size providing improved complexity estimates. 
They dub this metric the ``apparent complexity'', since it is not actually a bound on the original state's complexity, and is arbitrary in the choice and degree of coarse-graining. 
We adapt their method and apply it to neural networks based on a central insight: \textbf{the loss function induces a coarse-graining}, which can be understood as principled \textit{lossy} compression of the model.

\begin{figure}
    \centering
    \includegraphics[width=0.75\linewidth]{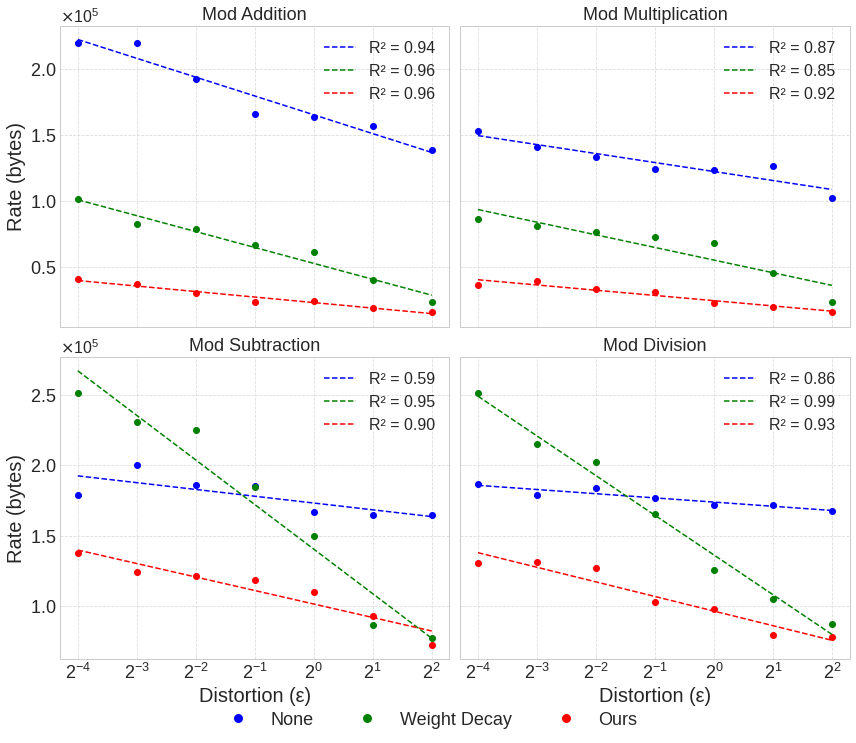}
    \caption{Algorithmic rate--distortion curves $r_\theta(\hat{\theta})$ for all datasets at varying distortion bounds $\epsilon$. The rate is the complexity of the coarse-grained weights $\hat{\theta}$ at the end of training. Mean over six seeds.}
    \label{fig:distortionbound}
\end{figure}

\subsection{Lossy Compression}
Shannon's theorem provides a lower-bound on coding rates which can \textit{losslessly} compress signals, but more efficient coding rates are possible if some distortion is allowed. 
For example, JPEG compression removes high-frequency details from images which are perceptually irrelevant to the human eye, incurring some distortion. 
Similarly, we want to find the least detailed model which produces acceptable performance on the training data. 
Since neural networks are initialized randomly and trained via stochastic optimization, much of their parameter's information capacity stores noise. 
In algorithmic information theory, noise is precisely that which is incompressible, so if we na\"ively compress model weights, the resultant complexity estimate from the compressed file size is dominated by the noise stored in the weights. 
To overcome this, we develop a lossy compression scheme which lets us formalize the trade-off between information capacity in the model and the performance on the training data as measured by the loss function. 
Rate--distortion theory formalizes the trade-off between coding rate, induced distortion, and perceptual quality \citep{Blau2019}, though it is typically defined over random variables. 
\cite{kdistortion} generalize the theory of rate distortion to the algorithmic information setting. They define the algorithmic rate--distortion function $r_x(y)$, which is the minimum number of bits needed to specify $y$ with distortion relative to $x$ no more than $\epsilon$:
\begin{equation}\label{ratedistortion}
    r_x(y)=\min\{K(y): d(x,y)<\epsilon\}
\end{equation}
Where $d$ is some distortion function. 
We will take the ``input string'' $x$ to be the model parameter vector $\theta$, and the ``output'' $y$ to be the coarse-grained parameter vector $\hat{\theta}$. 
The choice of distortion function $d$ is critical. 
In particular, we are not interested in the distortion in parameter space, $|\theta - \hat{\theta}|^2$. 
Instead, we are interested in the distortion \textit{under the loss function}. 
Hence, we set:
\begin{equation}
    d(\theta,\hat{\theta}) = \left| \mathcal{L}(\theta, D) - \mathcal{L}(\hat{\theta},D)\right|
\end{equation}
Substituting this distortion function into Eqn~\ref{ratedistortion}, we can see that the optimum can be interpreted as the least complex set of model weights $\hat{\theta}$ which induce distortion under the loss function no more than $\epsilon$. 
We control the rate--distortion trade-off through the parameter $\epsilon$, which leads to the following distortion criterion to accept a coarse-grained approximation:
\begin{equation}\label{epsilonbound}
    \left| \mathcal{L}(\theta, D) - \mathcal{L}(\hat{\theta},D)\right| < \epsilon
\end{equation}

\begin{figure}
    \centering
    \includegraphics[width=0.75\linewidth]{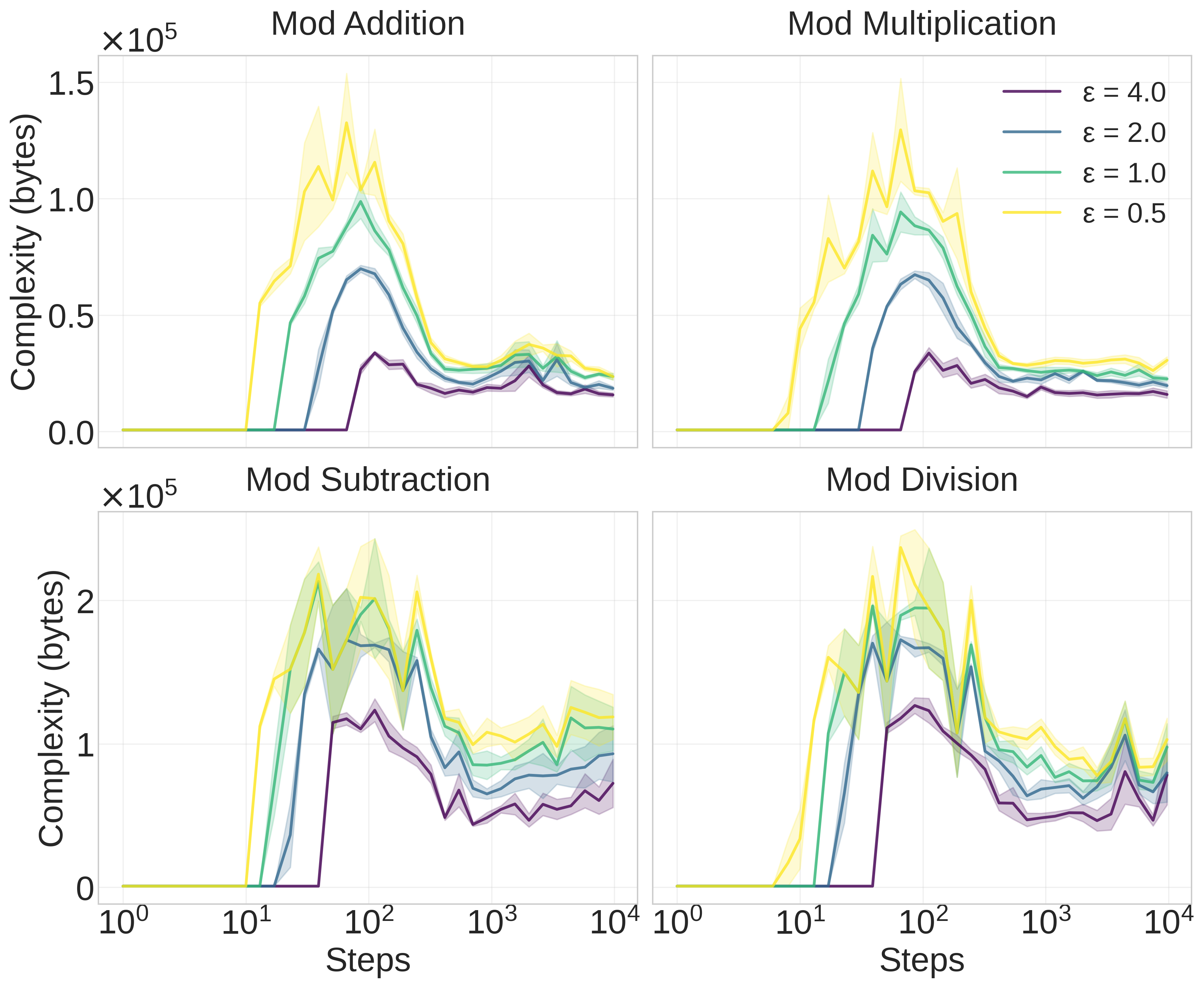}
    \caption{Complexity dynamics under varying distortion bounds $\epsilon$. As the allowed distortion increases, the estimated model complexity becomes smaller, as expected. The rise-and-fall pattern is robust under varying distortion bounds.}
    \label{complexitydynamicsvaryeps}
\end{figure}

Note that we could have chosen an alternative distortion function, such as difference in \textit{accuracy} or any other performance metric of interest. However, the cross-entropy loss preserves the interpretation of the model as a compressor of the training data. In fact, the MDL principle implies an optimal distortion bound $\epsilon$, which determines the best coarse-grained model $\hat{\theta}$:
\begin{equation}
    \epsilon \leftarrow \argmin_{\epsilon} H(D\mid \hat{\theta}) + r_\theta(\hat{\theta})
\end{equation}
Intuitively, as we coarse-grain a model it becomes less complex. 
However, it may also perform worse. 
The MDL principle implies that the optimally coarse-grained model is one which balances the complexity--performance trade-off, as measured by the model's ability to compress its training data, which is given by the loss function. 
In particular, a \textit{lossy model} might be a better \textit{lossless compressor} of the data, as measured by the total description length.
We approximate the algorithmic rate--distortion function via compression upper bounds, and plot curves of best fit in Fig~\ref{fig:distortionbound}. 

To obtain models which are optimal compressors, we must regularize them to limit their complexity. 
In the next section, we discuss the relationship between complexity, capacity, and regularization, and propose a regularization method which results in highly compressible models.

\section{Capacity and Regularization}
To control model complexity, practitioners introduce regularizers to the network training process. Most regularizers work by limiting a model's effective capacity, which \textit{may} limit its complexity. To see this, we consider a simplified representation of a model parameter. The information capacity $I$ of a single positive parameter with fixed precision $\delta$, and max size $\lambda$ is given by:
\begin{equation}\label{capacity}
    I = \log\frac{\lambda}{\delta}
\end{equation}
Under this simplified model, it is clear that we can limit the information capacity of the parameters either by controlling their range through $\lambda$, or by reducing their precision through $\delta$, which can be thought of as an effective quantization level. Regularizers such as weight decay control the information capacity by limiting $\lambda$, or proxies of it such as an $L^p$ norm. In particular, weight decay adds an independent $L^2$ penalty for each parameter to the loss function. Methods which control information capacity through $\delta$ include training in reduced precision \citep{mixedprecision}, and quantization-aware training \citep{1bit}.

However, training with regularization does not guarantee that the resultant model will be of low complexity. For example, while weight decay ensures parameter norms are smaller than they otherwise would have been, the network can still store detail in high-frequency information. Our experiments demonstrate that this indeed occurs in practice. A key requirement for complexity reduction therefore is a regularization scheme which explicitly alleviates this issue. 

A further complication is information which is spread across multiple parameters. Networks are typically parameterized by matrices which represent linear transformations, such that relevant information exists in their \textit{spectrum}. Ultimately such information is accounted for by the total sum of parameter capacities, but it is helpful to think of information in the spectral representation. We now introduce a regularization scheme which addresses both of these challenges. 


\subsection{Quantization and Noise}\label{noise}
In contrast to methods like weight decay, which regularize the capacity of networks by controlling their range through effective penalties on $\lambda$, we introduce a simple modification to the training procedure which has previously been studied by \cite{Hinton1993}, amongst many others: One way to set an effective lower bound on the quantization level $\delta$ from Eqn. \ref{capacity} is to train networks with noisy weights. The noise prevents the parameters from forming structure below the noise threshold, limiting the information capacity ``from the bottom''. To do this, we construct a set of noisy weights $\tilde{\theta}$ by adding independent Gaussian noise scaled by $\delta$ to every parameter at each forward pass:
\begin{equation}\label{noisyweights}
    \tilde{\theta} = \theta + \delta \mathcal{N}(0,1)
\end{equation}
We compute the forward pass and gradients with the noisy weights, then apply those gradients to the original parameters. 
Note that $\delta$ in Eqn~\ref{noisyweights} need not be the same as $\delta$ from Eqn.~\ref{capacity}, but applying this training procedure sets an effective lower bound on $\delta$ from Eqn.~\ref{capacity} (the effective precision of our parameters) so we use the same symbol.  
This implies that we ought to be able to remove all information below the scale set by $\delta$, i.e. that we can quantize the parameters with bin spacing at least $\delta$. 
To do this, we introduce a quantization operator $\hat{Q}$ which takes a parameter vector $\theta$ and bin size $\Delta$, and rounds all parameters to the nearest bin value
\begin{equation}
    \hat{Q}(\theta,\Delta) = \round{\frac{\theta}{\Delta}}\times\Delta
\end{equation}
We could achieve greater compression by using a separate $\Delta$ for each parameter group, but for simplicity we use a global value of $\Delta$. 
A simple algorithm which estimates the complexity of a network based on quantization coarse-graining alone is given in Algorithm~\ref{alg:complexity} in the appendix.

\subsection{Spectral Entropy}\label{entropysection}
While coarse-graining by quantization is simple and effective, it can only address high-frequency noise. 
In this section we introduce another coarse-graining procedure which takes advantage of low-rank approximation to remove random information. 
We use this low-rank approximation both as a regularization method to limit model complexity, and as a coarse-graining method during the compression stage, to produce tighter complexity estimates throughout training.

Low rank approximation is the problem of finding an approximating matrix $\tilde{M}$ for a matrix $M$ subject to the constraint that $\text{rank}(\tilde{M}) \leq r$ for some $r$, typically $r \ll \text{rank}(M)$. Low-rank approximation of neural network weight matrices has recently been used to achieve compute-- and parameter--efficient finetuning of LLMs \citep{lora, qlora}, building on the insight that both the weights and update gradients live on a low-dimension subspace of the total weight space \citep{intrinsicdimension, intrinsicdimensionupdate}. 

The mismatch between the apparent and effective ranks of the largest models is another manifestation of the difference between the capacity and complexity of networks. 
While large capacity may be needed for training dynamics to find optimal solutions, those learned solutions which generalize best tend to be of low effective rank, and therefore low complexity.

It is well-known that the optimal rank $k$ approximation of $M$ under the Frobenius norm is obtained by computing the singular value decomposition of $M$, and removing all but the $k$ largest singular values. Recall that for an $m\times n$ matrix $M \in \mathbb{R}^{m\times n}$, the singular value decomposition is $M = USV = \sum_{i=1}^r \sigma_i u_i v_i^{\intercal}$, where the columns of $U$ and the rows of $V$ are the left- and right-singular vectors respectively, and $S$ is a diagonal matrix $S=\text{diag}(\sigma_i)$ with singular values $\sigma_1 > \sigma_2 >...>\sigma_{r=\min(m,n)}$ in decreasing order. Then the optimal rank $k$ approximation is $\tilde{M} = \sum_{i=1}^k \sigma_i u_i v_i^{\intercal}$.

To coarse-grain, we search for the smallest rank decomposition of each layer that satisfies the $\epsilon$--bound of Eqn.~\ref{epsilonbound}. Note that different layers may admit different degrees of rank decomposition, so using a global value of $k$ across all layers of the network would be suboptimal. Instead, we search for optimal per-layer low-rank approximations by normalizing the singular values
\begin{equation}
    \bar\sigma_i = \frac{\sigma_i}{\sum_j\sigma_j} 
\end{equation}
so that $\sum_i\bar\sigma_i = 1$. Next we define a parameter $\tau \in (0,1)$ which sets an approximation threshold, so that the sum of the normalized singular values is no less than $\tau$. Intuitively, $\tau$ is a parameter which allows for convenient control of the approximation quality of the low-rank decomposition. For example, a matrix with only one large singular value, and many other negligibly small singular values has an effective rank of 1, so we will probably be able to satisfy the $\epsilon$--bound with a rank 1 approximation of that matrix, whereas for a matrix with many singular values of similar magnitude, our approximation must be close to full rank. To represent each layer with as few parameters as possible, we define a per-layer truncation threshold $k(\tau)$:
\begin{equation}
    k(\tau) = \min\{k : \sum_{i=1}^k \bar\sigma_{i} \geq \tau \}
\end{equation}
where the same threshold $\tau$ is used across all weight matrices. This lets us define a low-rank approximation operator $\tilde{R}(\theta, \tau)$ which returns the truncated singular value decomposition with effective rank controlled by $\tau$. Note that if the effective rank of the matrix is large, there is no compression benefit to the spectral representation. An $m\times n$ matrix $M$ normally requires $mn$ parameters for representation, and the SVD of rank $k$ uses $k(m+n)$ parameters, so the low-rank approximation is only useful if $k(\tau) < \frac{mn}{m+n}$.

\begin{equation}
\tilde{R}(\theta_j,\tau) = \begin{cases}
\left(\sigma^j_{i:k(\tau)}, u^j_{i:k(\tau)}, v^j_{i:k(\tau)}\right) & \text{if } k(\tau) < \frac{mn}{m+n} \\
\theta_j &  \text{otherwise}
\end{cases}
\end{equation}
Where $\theta_j$ is the network parameter group of layer $j$, and $(\sigma^j_{i:k(\tau)}, u^j_{i:k(\tau)}, v^j_{i:k(\tau)})$ are the corresponding low-rank matrices of rank $k(\tau)$. 
Note that we can interpret the normalized singular values as a probability distribution. Then, define the \textit{spectral entropy} of a matrix $H_{\text{svd}}$,
\begin{equation}\label{spectralentropy}
    H_{\text{svd}} = -\sum_i\bar\sigma_i\log\bar\sigma_i
\end{equation}
Intuitively, $H_{\text{svd}}$ tells us about the effective rank of the matrix by measuring the spread of the singular values. To see that, note that if all singular values are equal, $H_{\text{svd}}$ is maximized, and if all but one of the singular values are zero, $H_{\text{svd}}$ is minimized. \citet{Roy2007} formalize this by proving that $H_\text{svd}$ computes the effective rank of a matrix, $\text{rank}_\text{eff}(M) = \exp(H_\text{svd}(M))$, which measures the intrinsic dimension of the transformation parameterized by $M$. The spectral entropy not only provides a means of measuring the effective dimension of networks as training progresses---we can also explicitly penalize this quantity to encourage the network to learn simple, low-rank representations. In experiments, we show that our regularization method causes grokking, and produces the lowest-rank and most compressible networks of the methods we study.

\begin{figure}
    \centering
    \includegraphics[width=0.75\linewidth]{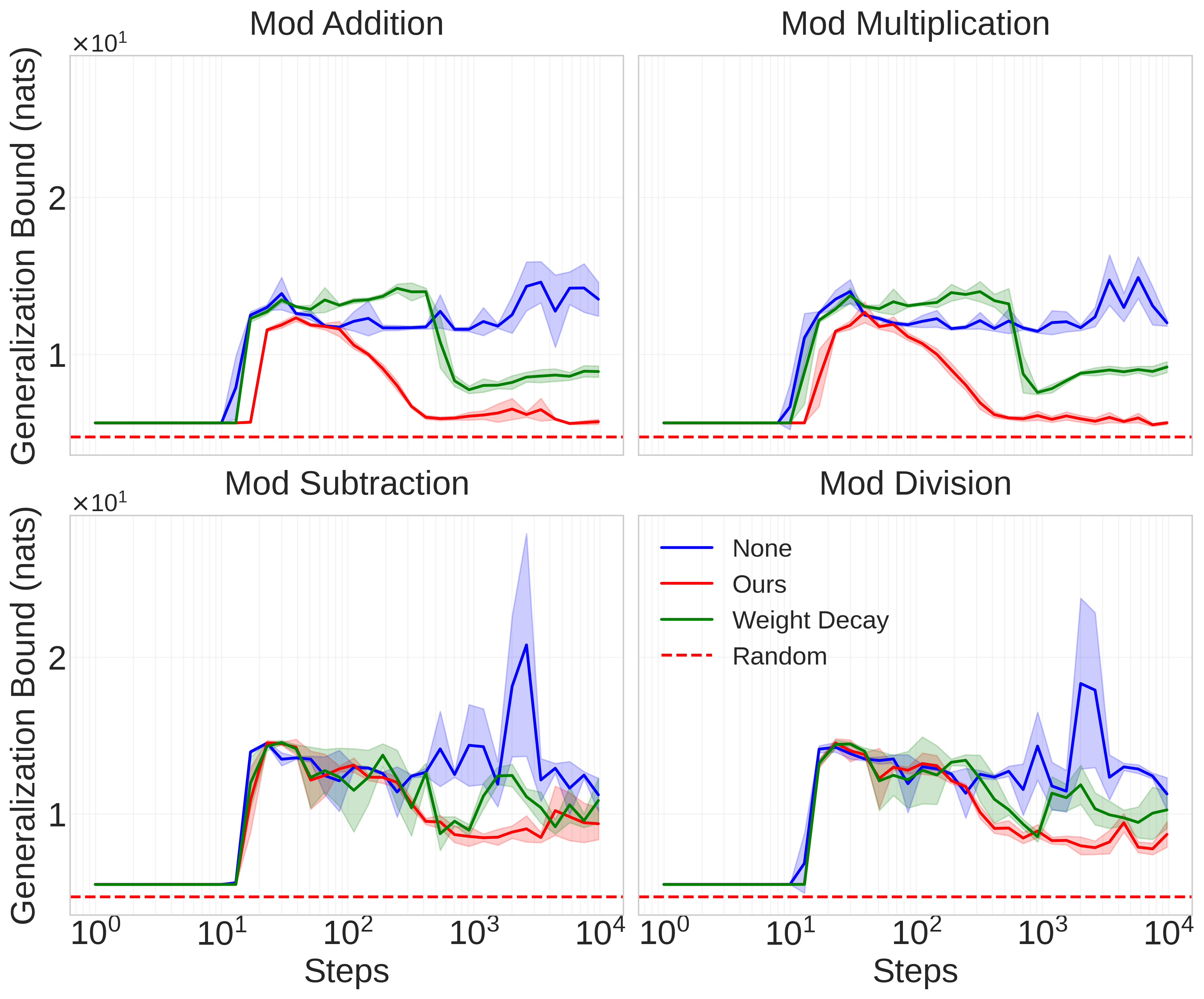}
    \caption{We plot the generalization bound on the expected risk according to Eqn. \ref{genbound}. While no models achieve low enough complexity to give non-vacuous bounds, our regularization method achieves the best bounds in all cases. It is not surprising that we can only achieve a vacuous bound, since the models are vastly over-parameterized, and the dataset small.}
    \label{genboundplot}
\end{figure}

Our full coarse-graining procedure is to perform a Bayesian optimization for the values of $\tau$ and $\Delta$ which produce the most compressed description, while satisfying the $\epsilon$--bound. When the low-rank description is more compressible, we quantize the \textit{decomposed} matrices. The full procedure is given in Algorithm~\ref{alg:complexity2}. We emphasize that the low-rank decomposition is used for two separate purposes: 1) the spectral entropy penalty is used as a regularizer, and 2) the low-rank decomposition is used to obtain compressed descriptions of the model at the coarse-graining stage.

The regularizer we study in experiments adds the spectral entropy penalty scaled by $\beta$ to the loss,
\begin{equation}
    \mathcal{L}_{\text{reg}}(\theta) = \beta H_{\text{svd}}(\theta)
\end{equation}
along with the noisy gradients described in section \ref{noise}, plus weight decay via the AdamW optimizer.

\begin{algorithm}\label{alg:complexity2}
\SetAlgoLined
\KwIn{Neural network parameters $\theta$, number of optimization steps $N$, error tolerance $\epsilon$}
\KwOut{Optimal compressed size}
\caption{Bayesian Optimization for Compression by Coarse-Graining}
Initialize Bayesian Optimizer $BO$\;
best\_compressed\_size $\gets \infty$\;
\For{$i = 1$ \KwTo $N$}{
    $\tau_i, \Delta_i \gets BO.\textsc{SuggestParameters}()$\;
    $\theta_R \gets R(\theta, \tau_i)$ \tcp*{Low-rank approximation}
    $\theta_{RQ} \gets Q(\theta_R, \Delta_i)$ \tcp*{Quantization}
    \eIf{$\left|\textsc{Loss}(\theta_{RQ}) - \textsc{Loss}(\theta)\right| < \epsilon$}{
        compressed\_size $\gets \textsc{compress}(\theta_{RQ})$\;
        \If{compressed\_size $<$ best\_compressed\_size}{
            best\_compressed\_size $\gets$ compressed\_size\;
        }
        $BO.\textsc{UpdateModel}(\tau_i, \Delta_i, \text{compressed\_size})$\;
    }{
        $BO.\textsc{UpdateModel}(\tau_i, \Delta_i, \infty)$ \tcp*{Penalize invalid solutions}
    }
}
\Return best\_compressed\_size
\end{algorithm}

\section{Experiments}\label{experiments}

To study the complexity dynamics of networks transitioning from memorizing to generalizing solutions, we adopt the ``grokking'' tasks first reported in \cite{grokking}. These are a set of modular arithmetic tasks which networks easily over-fit, quickly achieving zero training error while test error remains large, suggesting memorization of the training examples. After many further steps of gradient descent after over-fitting, networks suddenly generalize and achieve perfect test accuracy. Although grokking is a sign of improperly regularized training, the extended delay between memorization and generalization is convenient for understanding complexity dynamics, and their relationship with generalization.

We demonstrate a regularizer, $\mathcal{L}_{\text{reg}}$ on the grokking tasks and show that 1) it causes grokking 2) it results in lower complexity networks than weight decay alone, the most widely-used regularization method, as measured by our complexity metric.

We reproduce a subset of the grokking experiments first reported in \cite{grokking}. Our training tasks all consist of learning a binary operation mod a prime number $p=113$. For each binary operation we construct a dataset of equations of the form 
$\langle x \rangle \langle \text{op} \rangle \langle y \rangle \langle = \rangle \langle x \circ y \rangle$, where $\langle a \rangle$ stands for the token corresponding to element $a$. We train standard decoder-only transformers with a single layer for mod addition and multiplication, and two layers for mod subtraction and division. Further training details can be found in the appendix. We use Algorithm~\ref{alg:complexity2} to compute complexity upper-bounds on the coarse-grained models. The final compression step is performed by \texttt{bzip2}, an off-the-shelf compression utility. 

First, we establish that grokking occurs in all regularized models, and does not occur in unregularized models in Appendix Figure \ref{accuracy}. Next, we plot accuracy and complexity together for the regularized models in Fig~\ref{complexaccuracy}, which shows our main result: as expected from Occam's Razor, complexity is maximized during memorization and falls during generalization, demonstrating a clear complexity phase transition. In contrast, Fig~\ref{unregularizedcomplexity} demonstrates that without regularization, models tend to stay in the high-complexity memorization regime indefinitely. Next, we plot algorithmic rate--distortion curves of best fit in Fig~\ref{fig:distortionbound}. We fit curves of the form $y=a\log(x)+b$, and display $R^2$ values in the legend. These curves show how lossy \textit{model} compression trades off with lossless \textit{data} compression for the converged models at varying distortion bounds. Fig~\ref{complexitydynamicsvaryeps} shows how the complexity dynamics change under varying distortion levels $\epsilon$: we see that as the distortion bound is loosened (larger $\epsilon$), the corresponding complexity falls for the same model, as expected from the theory of rate--distortion.

Plotting the generalization bound from Eqn.~\ref{genbound}, we see in Fig~\ref{genboundplot} the models trained with our regularizer come closest to a non-trivial generalization bound: they achieve the least sum of data entropy and model complexity. Fig~\ref{TDL} in the appendix plots the total description length of the dataset under each model, which mirrors the generalization bound.

\section{Conclusion}

Just as physical systems can transition between distinct phases of matter, we show that learning systems can transition between distinct phases characterized by the complexity of their internal representations. We suggest that a complexity phase transition is the central mechanism driving the grokking phenomenon, in contrast to prior works which focus on more specific signatures such as weight norms \citep{varma2023} or linear structure \citep{liu2022}. We justify our analysis on theoretical grounds by linking complexity to generalization through formal statistical generalization bounds \citep{Lotfi2024}. From this perspective, we can see that small weight norms or linear structure are different ways for networks to be simple, but these metrics alone are insufficient explanations since complexity can emerge through a variety of mechanisms. For example, we showed how the effective rank of the network is one means by which complexity can emerge, which we penalize through the spectral entropy measure we introduced in Section \ref{entropysection}.

The distinctive separation between the memorizing and generalizing phases makes the grokking phenomenon an ideal model system to initiate the study of complexity phase transitions in learning systems. A related pathological phenomenon observed in learning systems is \textit{double descent} \citep{deepdoubledescent}, in which test error first falls, then rises, then falls again as a function of model size, with training steps fixed. A unified theory of complexity and generalization should predict when and under which conditions such training dynamics occur, and may suggest avenues for further progress in the theory of machine learning. Generalizing our results to account for arbitrary learning systems could lead to a deeper understanding of the interplay between optimization dynamics, the loss landscape, data, model capacity, and complexity. 

Equipped with such a theory, practitioners could predict the performance of machine learning systems before running costly experiments, and insights gained from a more complete theory of machine learning could enable new breakthroughs in our understanding of complex systems more broadly. The emerging field of algorithmic thermodynamics \citep{Ebtekar2025, BAEZ_STAY_2012, Wolpert2020} provides a promising mathematical and computational framework to analyze these systems.

The study of phase transitions in learning systems provides a window into fundamental questions about the nature of generalization and the emergence of structure in dynamical systems. The rise and fall of complexity we report in this work is conjectured to be a generic property of certain complex dynamical systems \citep{aaronson2014}, and understanding its origin in greater detail could help explain not only the emergence of abstraction and generalization in learning systems, but also the apparent complexity we observe in the natural world more broadly. Machine learning systems such as neural networks are convenient model complex systems to study from a theoretical perspective due to their transparency and controllability, and the study of their complexity-theoretic properties is a promising avenue for further research.

\section*{Acknowledgments}
The authors thank Jasmine Brewer and Andrew P.~Turner for valuable discussions that strengthened this work.

\section*{Funding}
This work was supported by the EPSRC Programme Grant \emph{“From Sensing to Collaboration”} (EP/V000748/1). Branton DeMoss was supported by an AWS Lighthouse Scholarship.

\bibliographystyle{elsarticle-num-names}\biboptions{authoryear}
\bibliography{refs}  

\begin{thebibliography}{31}
\expandafter\ifx\csname natexlab\endcsname\relax\def\natexlab#1{#1}\fi
\providecommand{\url}[1]{\texttt{#1}}
\providecommand{\href}[2]{#2}
\providecommand{\path}[1]{#1}
\providecommand{\DOIprefix}{doi:}
\providecommand{\ArXivprefix}{arXiv:}
\providecommand{\URLprefix}{URL: }
\providecommand{\Pubmedprefix}{pmid:}
\providecommand{\doi}[1]{\href{http://dx.doi.org/#1}{\path{#1}}}
\providecommand{\Pubmed}[1]{\href{pmid:#1}{\path{#1}}}
\providecommand{\bibinfo}[2]{#2}
\ifx\xfnm\relax \def\xfnm[#1]{\unskip,\space#1}\fi
\bibitem[{Power et~al.(2022)Power, Burda, Edwards, Babuschkin, and
  Misra}]{grokking}
\bibinfo{author}{A.~Power}, \bibinfo{author}{Y.~Burda},
  \bibinfo{author}{H.~Edwards}, \bibinfo{author}{I.~Babuschkin},
  \bibinfo{author}{V.~Misra},
\newblock \bibinfo{title}{Grokking: Generalization beyond overfitting on small
  algorithmic datasets},
\newblock \bibinfo{journal}{ArXiv} \bibinfo{volume}{abs/2201.02177}
  (\bibinfo{year}{2022}).
\bibitem[{Aaronson et~al.(2014)Aaronson, Carroll, and Ouellette}]{aaronson2014}
\bibinfo{author}{S.~Aaronson}, \bibinfo{author}{S.~M. Carroll},
  \bibinfo{author}{L.~Ouellette}, \bibinfo{title}{Quantifying the rise and fall
  of complexity in closed systems: The coffee automaton}, \bibinfo{year}{2014}.
  \URLprefix \url{https://arxiv.org/abs/1405.6903}.
  \href{http://arxiv.org/abs/1405.6903}{{\tt arXiv:1405.6903}}.
\bibitem[{Rissanen(1978)}]{Rissanen1978}
\bibinfo{author}{J.~Rissanen},
\newblock \bibinfo{title}{Modeling by shortest data description*},
\newblock \bibinfo{journal}{Autom.} \bibinfo{volume}{14} (\bibinfo{year}{1978})
  \bibinfo{pages}{465--471}.
\bibitem[{Shannon(1948)}]{Shannon1948}
\bibinfo{author}{C.~E. Shannon},
\newblock \bibinfo{title}{A mathematical theory of communication},
\newblock \bibinfo{journal}{Bell Syst. Tech. J.} \bibinfo{volume}{27}
  (\bibinfo{year}{1948}) \bibinfo{pages}{623--656}.
\bibitem[{Vapnik(1991)}]{Vapnik1991}
\bibinfo{author}{V.~N. Vapnik},
\newblock \bibinfo{title}{Principles of risk minimization for learning theory},
\newblock in: \bibinfo{booktitle}{Neural Information Processing Systems},
  \bibinfo{year}{1991}.
\bibitem[{Bartlett and Mendelson(2003)}]{Rademacher}
\bibinfo{author}{P.~L. Bartlett}, \bibinfo{author}{S.~Mendelson},
\newblock \bibinfo{title}{Rademacher and gaussian complexities: Risk bounds and
  structural results},
\newblock \bibinfo{journal}{J. Mach. Learn. Res.} \bibinfo{volume}{3}
  (\bibinfo{year}{2003}) \bibinfo{pages}{463--482}.
\bibitem[{Vapnik and Chervonenkis(1971)}]{VCDimension}
\bibinfo{author}{V.~N. Vapnik}, \bibinfo{author}{A.~Chervonenkis},
\newblock \bibinfo{title}{On the uniform convergence of relative frequencies of
  events to their probabilities},
\newblock \bibinfo{year}{1971}.
\bibitem[{Liu et~al.(2022{\natexlab{a}})Liu, Kitouni, Nolte, Michaud, Tegmark,
  and Williams}]{liu2022}
\bibinfo{author}{Z.~Liu}, \bibinfo{author}{O.~Kitouni},
  \bibinfo{author}{N.~Nolte}, \bibinfo{author}{E.~J. Michaud},
  \bibinfo{author}{M.~Tegmark}, \bibinfo{author}{M.~Williams},
\newblock \bibinfo{title}{Towards understanding grokking: An effective theory
  of representation learning},
\newblock \bibinfo{journal}{ArXiv} \bibinfo{volume}{abs/2205.10343}
  (\bibinfo{year}{2022}{\natexlab{a}}).
\bibitem[{Liu et~al.(2022{\natexlab{b}})Liu, Michaud, and Tegmark}]{omnigrok}
\bibinfo{author}{Z.~Liu}, \bibinfo{author}{E.~J. Michaud},
  \bibinfo{author}{M.~Tegmark},
\newblock \bibinfo{title}{Omnigrok: Grokking beyond algorithmic data},
\newblock \bibinfo{journal}{ArXiv} \bibinfo{volume}{abs/2210.01117}
  (\bibinfo{year}{2022}{\natexlab{b}}).
\bibitem[{Varma et~al.(2023)Varma, Shah, Kenton, Kram'ar, and
  Kumar}]{varma2023}
\bibinfo{author}{V.~Varma}, \bibinfo{author}{R.~Shah},
  \bibinfo{author}{Z.~Kenton}, \bibinfo{author}{J.~Kram'ar},
  \bibinfo{author}{R.~Kumar},
\newblock \bibinfo{title}{Explaining grokking through circuit efficiency},
\newblock \bibinfo{journal}{ArXiv} \bibinfo{volume}{abs/2309.02390}
  (\bibinfo{year}{2023}).
\bibitem[{Nanda et~al.(2023)Nanda, Chan, Lieberum, Smith, and
  Steinhardt}]{Nanda2023}
\bibinfo{author}{N.~Nanda}, \bibinfo{author}{L.~Chan},
  \bibinfo{author}{T.~Lieberum}, \bibinfo{author}{J.~Smith},
  \bibinfo{author}{J.~Steinhardt},
\newblock \bibinfo{title}{Progress measures for grokking via mechanistic
  interpretability},
\newblock \bibinfo{journal}{ArXiv} \bibinfo{volume}{abs/2301.05217}
  (\bibinfo{year}{2023}).
\bibitem[{Nakkiran et~al.(2019)Nakkiran, Kaplun, Bansal, Yang, Barak, and
  Sutskever}]{deepdoubledescent}
\bibinfo{author}{P.~Nakkiran}, \bibinfo{author}{G.~Kaplun},
  \bibinfo{author}{Y.~Bansal}, \bibinfo{author}{T.~Yang},
  \bibinfo{author}{B.~Barak}, \bibinfo{author}{I.~Sutskever},
\newblock \bibinfo{title}{Deep double descent: where bigger models and more
  data hurt},
\newblock \bibinfo{journal}{Journal of Statistical Mechanics: Theory and
  Experiment} \bibinfo{volume}{2021} (\bibinfo{year}{2019}).
\bibitem[{Kaplan et~al.(2020)Kaplan, McCandlish, Henighan, Brown, Chess, Child,
  Gray, Radford, Wu, and Amodei}]{kaplan2020}
\bibinfo{author}{J.~Kaplan}, \bibinfo{author}{S.~McCandlish},
  \bibinfo{author}{T.~Henighan}, \bibinfo{author}{T.~B. Brown},
  \bibinfo{author}{B.~Chess}, \bibinfo{author}{R.~Child},
  \bibinfo{author}{S.~Gray}, \bibinfo{author}{A.~Radford},
  \bibinfo{author}{J.~Wu}, \bibinfo{author}{D.~Amodei}, \bibinfo{title}{Scaling
  laws for neural language models}, \bibinfo{year}{2020}. \URLprefix
  \url{https://arxiv.org/abs/2001.08361}.
  \href{http://arxiv.org/abs/2001.08361}{{\tt arXiv:2001.08361}}.
\bibitem[{Henighan et~al.(2020)Henighan, Kaplan, Katz, Chen, Hesse, Jackson,
  Jun, Brown, Dhariwal, Gray, Hallacy, Mann, Radford, Ramesh, Ryder, Ziegler,
  Schulman, Amodei, and McCandlish}]{henighan2020}
\bibinfo{author}{T.~Henighan}, \bibinfo{author}{J.~Kaplan},
  \bibinfo{author}{M.~Katz}, \bibinfo{author}{M.~Chen},
  \bibinfo{author}{C.~Hesse}, \bibinfo{author}{J.~Jackson},
  \bibinfo{author}{H.~Jun}, \bibinfo{author}{T.~B. Brown},
  \bibinfo{author}{P.~Dhariwal}, \bibinfo{author}{S.~Gray},
  \bibinfo{author}{C.~Hallacy}, \bibinfo{author}{B.~Mann},
  \bibinfo{author}{A.~Radford}, \bibinfo{author}{A.~Ramesh},
  \bibinfo{author}{N.~Ryder}, \bibinfo{author}{D.~M. Ziegler},
  \bibinfo{author}{J.~Schulman}, \bibinfo{author}{D.~Amodei},
  \bibinfo{author}{S.~McCandlish},
\newblock \bibinfo{title}{Scaling laws for autoregressive generative modeling},
\newblock \bibinfo{journal}{ArXiv} \bibinfo{volume}{abs/2010.14701}
  (\bibinfo{year}{2020}).
\bibitem[{Lotfi et~al.(2024)Lotfi, Finzi, Kuang, Rudner, Goldblum, and
  Wilson}]{Lotfi2024}
\bibinfo{author}{S.~Lotfi}, \bibinfo{author}{M.~Finzi},
  \bibinfo{author}{Y.~Kuang}, \bibinfo{author}{T.~G.~J. Rudner},
  \bibinfo{author}{M.~Goldblum}, \bibinfo{author}{A.~G. Wilson},
\newblock \bibinfo{title}{Non-vacuous generalization bounds for large language
  models},
\newblock \bibinfo{journal}{ICML 2024}  (\bibinfo{year}{2024}). \URLprefix
  \url{https://arxiv.org/abs/2312.17173}.
  \href{http://arxiv.org/abs/2312.17173}{{\tt arXiv:2312.17173}}.
\bibitem[{Goldblum et~al.(2023)Goldblum, Finzi, Rowan, and
  Wilson}]{Goldblum2023}
\bibinfo{author}{M.~Goldblum}, \bibinfo{author}{M.~Finzi},
  \bibinfo{author}{K.~Rowan}, \bibinfo{author}{A.~G. Wilson},
\newblock \bibinfo{title}{The no free lunch theorem, kolmogorov complexity, and
  the role of inductive biases in machine learning},
\newblock \bibinfo{journal}{ArXiv} \bibinfo{volume}{abs/2304.05366}
  (\bibinfo{year}{2023}).
\bibitem[{Langford and Seeger(2001)}]{finitehypothesis}
\bibinfo{author}{J.~Langford}, \bibinfo{author}{M.~Seeger},
\newblock \bibinfo{title}{Bounds for averaging classifiers.},
\newblock \bibinfo{journal}{Carnegie Mellon}  (\bibinfo{year}{2001}).
  \URLprefix
  \url{https://www.cs.cmu.edu/~jcl/papers/averaging/averaging_tech.pdf}.
\bibitem[{Li and Vit{\'a}nyi(1993)}]{LiVitanyi}
\bibinfo{author}{M.~Li}, \bibinfo{author}{P.~M.~B. Vit{\'a}nyi},
\newblock \bibinfo{title}{An introduction to kolmogorov complexity and its
  applications},
\newblock in: \bibinfo{booktitle}{Graduate Texts in Computer Science},
  \bibinfo{year}{1993}.
\bibitem[{Blau and Michaeli(2019)}]{Blau2019}
\bibinfo{author}{Y.~Blau}, \bibinfo{author}{T.~Michaeli},
\newblock \bibinfo{title}{Rethinking lossy compression: The
  rate-distortion-perception tradeoff},
\newblock in: \bibinfo{booktitle}{International Conference on Machine
  Learning}, \bibinfo{year}{2019}.
\bibitem[{Vereshchagin and Vit{\'a}nyi(2004)}]{kdistortion}
\bibinfo{author}{N.~K. Vereshchagin}, \bibinfo{author}{P.~M.~B. Vit{\'a}nyi},
\newblock \bibinfo{title}{Rate distortion and denoising of individual data
  using kolmogorov complexity},
\newblock \bibinfo{journal}{IEEE Transactions on Information Theory}
  \bibinfo{volume}{56} (\bibinfo{year}{2004}) \bibinfo{pages}{3438--3454}.
\bibitem[{Micikevicius et~al.(2017)Micikevicius, Narang, Alben, Diamos, Elsen,
  Garc{\'i}a, Ginsburg, Houston, Kuchaiev, Venkatesh, and Wu}]{mixedprecision}
\bibinfo{author}{P.~Micikevicius}, \bibinfo{author}{S.~Narang},
  \bibinfo{author}{J.~Alben}, \bibinfo{author}{G.~F. Diamos},
  \bibinfo{author}{E.~Elsen}, \bibinfo{author}{D.~Garc{\'i}a},
  \bibinfo{author}{B.~Ginsburg}, \bibinfo{author}{M.~Houston},
  \bibinfo{author}{O.~Kuchaiev}, \bibinfo{author}{G.~Venkatesh},
  \bibinfo{author}{H.~Wu},
\newblock \bibinfo{title}{Mixed precision training},
\newblock \bibinfo{journal}{ArXiv} \bibinfo{volume}{abs/1710.03740}
  (\bibinfo{year}{2017}).
\bibitem[{Ma et~al.(2024)Ma, Wang, Ma, Wang, Wang, Huang, Dong, Wang, Xue, and
  Wei}]{1bit}
\bibinfo{author}{S.~Ma}, \bibinfo{author}{H.~Wang}, \bibinfo{author}{L.~Ma},
  \bibinfo{author}{L.~Wang}, \bibinfo{author}{W.~Wang},
  \bibinfo{author}{S.~Huang}, \bibinfo{author}{L.~Dong},
  \bibinfo{author}{R.~Wang}, \bibinfo{author}{J.~Xue},
  \bibinfo{author}{F.~Wei},
\newblock \bibinfo{title}{The era of 1-bit llms: All large language models are
  in 1.58 bits},
\newblock \bibinfo{journal}{ArXiv} \bibinfo{volume}{abs/2402.17764}
  (\bibinfo{year}{2024}).
\bibitem[{Hinton and van Camp(1993)}]{Hinton1993}
\bibinfo{author}{G.~E. Hinton}, \bibinfo{author}{D.~van Camp},
\newblock \bibinfo{title}{Keeping the neural networks simple by minimizing the
  description length of the weights},
\newblock in: \bibinfo{booktitle}{Annual Conference Computational Learning
  Theory}, \bibinfo{year}{1993}.
\bibitem[{Hu et~al.(2021)Hu, Shen, Wallis, Allen-Zhu, Li, Wang, and
  Chen}]{lora}
\bibinfo{author}{J.~E. Hu}, \bibinfo{author}{Y.~Shen},
  \bibinfo{author}{P.~Wallis}, \bibinfo{author}{Z.~Allen-Zhu},
  \bibinfo{author}{Y.~Li}, \bibinfo{author}{S.~Wang},
  \bibinfo{author}{W.~Chen},
\newblock \bibinfo{title}{Lora: Low-rank adaptation of large language models},
\newblock \bibinfo{journal}{ArXiv} \bibinfo{volume}{abs/2106.09685}
  (\bibinfo{year}{2021}).
\bibitem[{Dettmers et~al.(2023)Dettmers, Pagnoni, Holtzman, and
  Zettlemoyer}]{qlora}
\bibinfo{author}{T.~Dettmers}, \bibinfo{author}{A.~Pagnoni},
  \bibinfo{author}{A.~Holtzman}, \bibinfo{author}{L.~Zettlemoyer},
\newblock \bibinfo{title}{Qlora: Efficient finetuning of quantized llms},
\newblock \bibinfo{journal}{ArXiv} \bibinfo{volume}{abs/2305.14314}
  (\bibinfo{year}{2023}).
\bibitem[{Li et~al.(2018)Li, Farkhoor, Liu, and Yosinski}]{intrinsicdimension}
\bibinfo{author}{C.~Li}, \bibinfo{author}{H.~Farkhoor},
  \bibinfo{author}{R.~Liu}, \bibinfo{author}{J.~Yosinski},
\newblock \bibinfo{title}{Measuring the intrinsic dimension of objective
  landscapes},
\newblock \bibinfo{journal}{ArXiv} \bibinfo{volume}{abs/1804.08838}
  (\bibinfo{year}{2018}).
\bibitem[{Aghajanyan et~al.(2020)Aghajanyan, Zettlemoyer, and
  Gupta}]{intrinsicdimensionupdate}
\bibinfo{author}{A.~Aghajanyan}, \bibinfo{author}{L.~Zettlemoyer},
  \bibinfo{author}{S.~Gupta},
\newblock \bibinfo{title}{Intrinsic dimensionality explains the effectiveness
  of language model fine-tuning},
\newblock \bibinfo{journal}{ArXiv} \bibinfo{volume}{abs/2012.13255}
  (\bibinfo{year}{2020}).
\bibitem[{Roy and Vetterli(2007)}]{Roy2007}
\bibinfo{author}{O.~Roy}, \bibinfo{author}{M.~Vetterli},
\newblock \bibinfo{title}{The effective rank: A measure of effective
  dimensionality},
\newblock \bibinfo{journal}{2007 15th European Signal Processing Conference}
  (\bibinfo{year}{2007}) \bibinfo{pages}{606--610}.
\bibitem[{Ebtekar and Hutter(2025)}]{Ebtekar2025}
\bibinfo{author}{A.~Ebtekar}, \bibinfo{author}{M.~Hutter},
\newblock \bibinfo{title}{Foundations of algorithmic thermodynamics},
\newblock \bibinfo{journal}{Phys. Rev. E} \bibinfo{volume}{111}
  (\bibinfo{year}{2025}) \bibinfo{pages}{014118}. \URLprefix
  \url{https://link.aps.org/doi/10.1103/PhysRevE.111.014118}.
  \DOIprefix\doi{10.1103/PhysRevE.111.014118}.
\bibitem[{Baez and Stay(2012)}]{BAEZ_STAY_2012}
\bibinfo{author}{J.~Baez}, \bibinfo{author}{M.~Stay},
\newblock \bibinfo{title}{Algorithmic thermodynamics},
\newblock \bibinfo{journal}{Mathematical Structures in Computer Science}
  \bibinfo{volume}{22} (\bibinfo{year}{2012}) \bibinfo{pages}{771–787}.
  \DOIprefix\doi{10.1017/S0960129511000521}.
\bibitem[{Kolchinsky and Wolpert(2020)}]{Wolpert2020}
\bibinfo{author}{A.~Kolchinsky}, \bibinfo{author}{D.~H. Wolpert},
\newblock \bibinfo{title}{Thermodynamic costs of turing machines},
\newblock \bibinfo{journal}{Phys. Rev. Res.} \bibinfo{volume}{2}
  (\bibinfo{year}{2020}) \bibinfo{pages}{033312}. \URLprefix
  \url{https://link.aps.org/doi/10.1103/PhysRevResearch.2.033312}.
  \DOIprefix\doi{10.1103/PhysRevResearch.2.033312}.

\end{thebibliography}

\appendix
\section{Appendix}

\begin{algorithm}
\SetAlgoLined
\KwData{params, epsilon}
\KwResult{best\_complexity}
\caption{Complexity by Quantization}
best\_complexity $\gets$ \textsc{compress}(params)\;
\For{bin\_sizing \textbf{from} small \textbf{to} large}{
    quantized\_params $\gets$ \textsc{Quantize}(params, bin\_sizing)\;
    \If{$\left|\textsc{Loss}(\text{quantized\_params}) - \textsc{Loss}(\text{params})\right| < \epsilon$}{
        best\_complexity $\gets$ \textsc{min}(\textsc{compress}(quantized\_params), best\_complexity)\;
    }
}
\Return best\_complexity
\label{alg:complexity}
\end{algorithm}


\begin{figure}[h]
    \centering
    \includegraphics[width=0.9\linewidth]{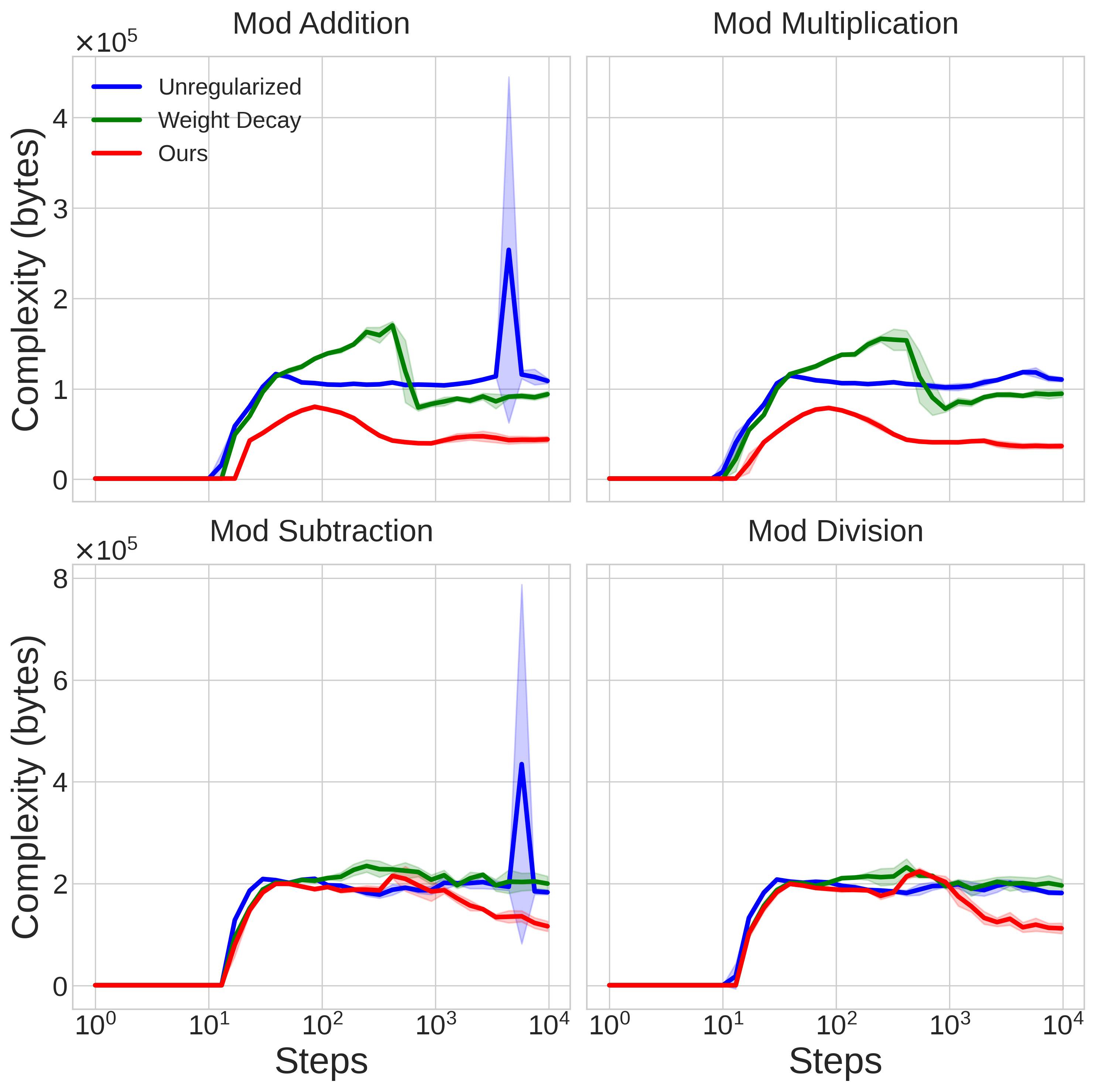}
    \caption{Complexity estimates using quantization coarse-graining alone, according to Algorithm \ref{alg:complexity}. Detecting the complexity dynamics requires relatively tight complexity bounds. While we see a similar rise and fall pattern in the regularized models of the top row, this method fails to distinguish these dynamics in the weight decayed models in the subtraction and addition. Note that the complexity bounds are about two times worse than those achieved by the spectral method in Fig \ref{complexaccuracy}.}
    \label{quantizationcomplexity}
\end{figure}

\begin{figure}[h]
    \centering
    \includegraphics[width=0.9\linewidth]{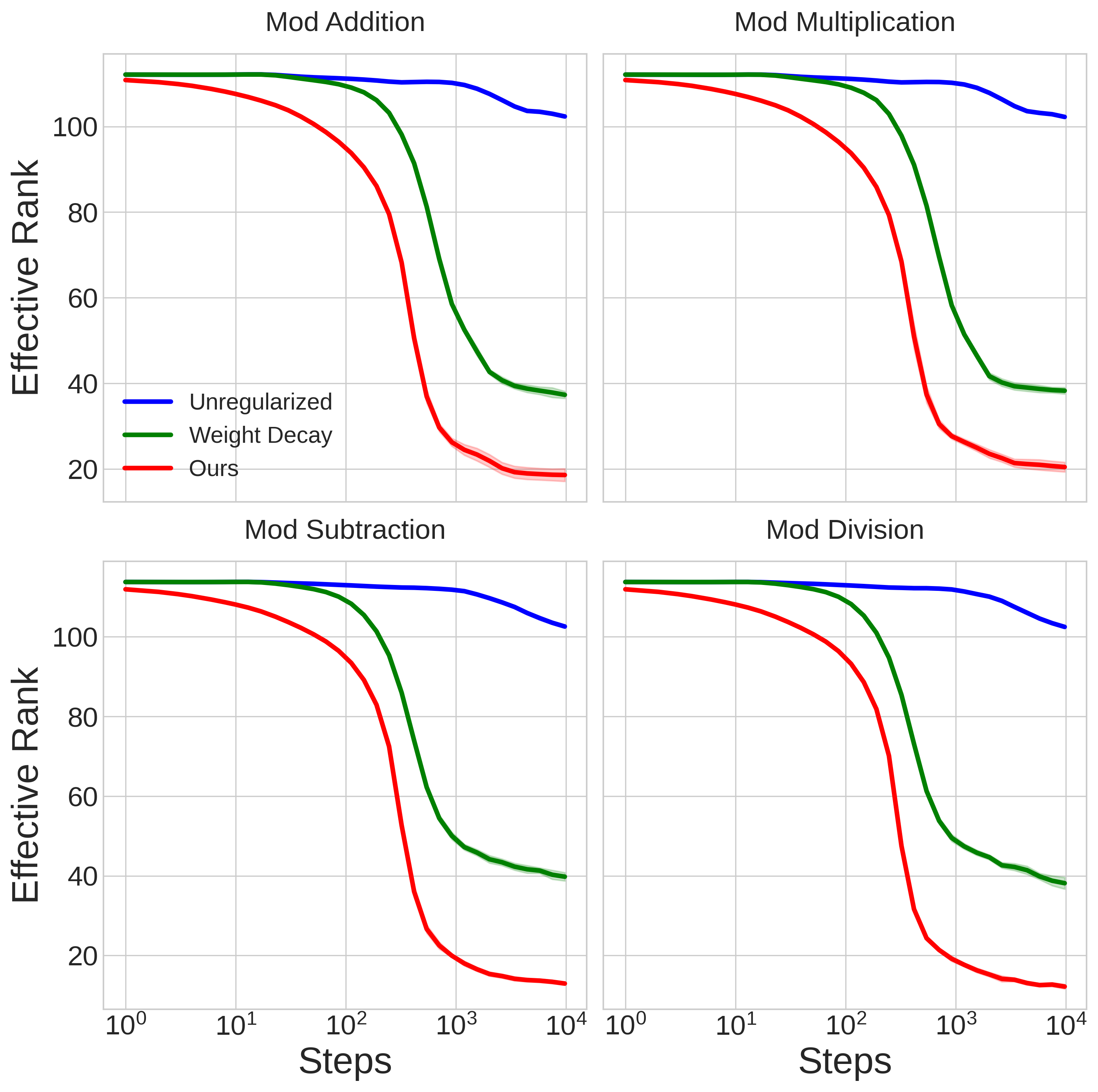}
    \caption{Effective rank ($\text{rank}_{\text{eff}} = e^{H_{\text{svd}}}$) of models throughout training. We see that the regularized models fall in effective rank when generalizing, while the unregularized model remains near full rank throughout training. Our regularizer explicitly penalizes this measure, and results in both the lowest rank and most compressible model amongst those we study.}
    \label{effectiverank}
\end{figure}

\begin{figure}[h]
    \centering
    \includegraphics[width=0.9\linewidth]{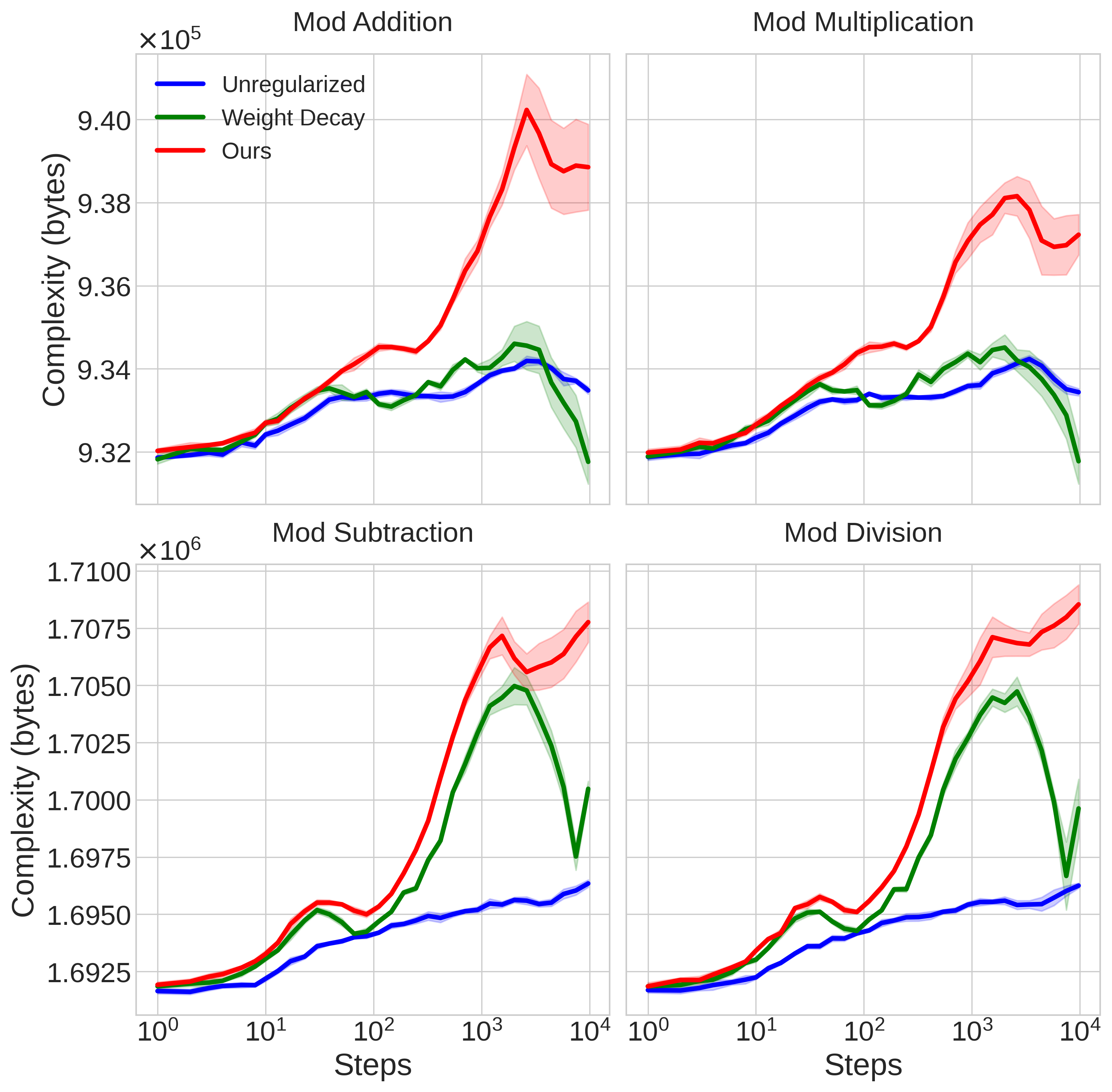}
    \caption{In this figure, we simply plot the na\"ive \texttt{bzip2} compression of the model weights without any coarse-graining (note the y-axis scales). The range of variation in these complexity estimates is very small throughout training---if we plotted it on the same scale as other complexity plots, it would simply look like a flat line from the starting estimate. The dynamics here may correspond to weight patterns induced by the different regularizers that the compression algorithm is particularly suited or unsuited to compressing. Intriguingly, the dips in the complexity bounds of the regularized models around $10^3$ steps corresponds to generalization, giving some slight hint that some change has occurred in the model. To fully reveal this phenomenon, the coarse-graining procedure is required.}
    \label{naivecompression}
\end{figure}

\begin{figure}[h]
    \centering
    \includegraphics[width=\linewidth]{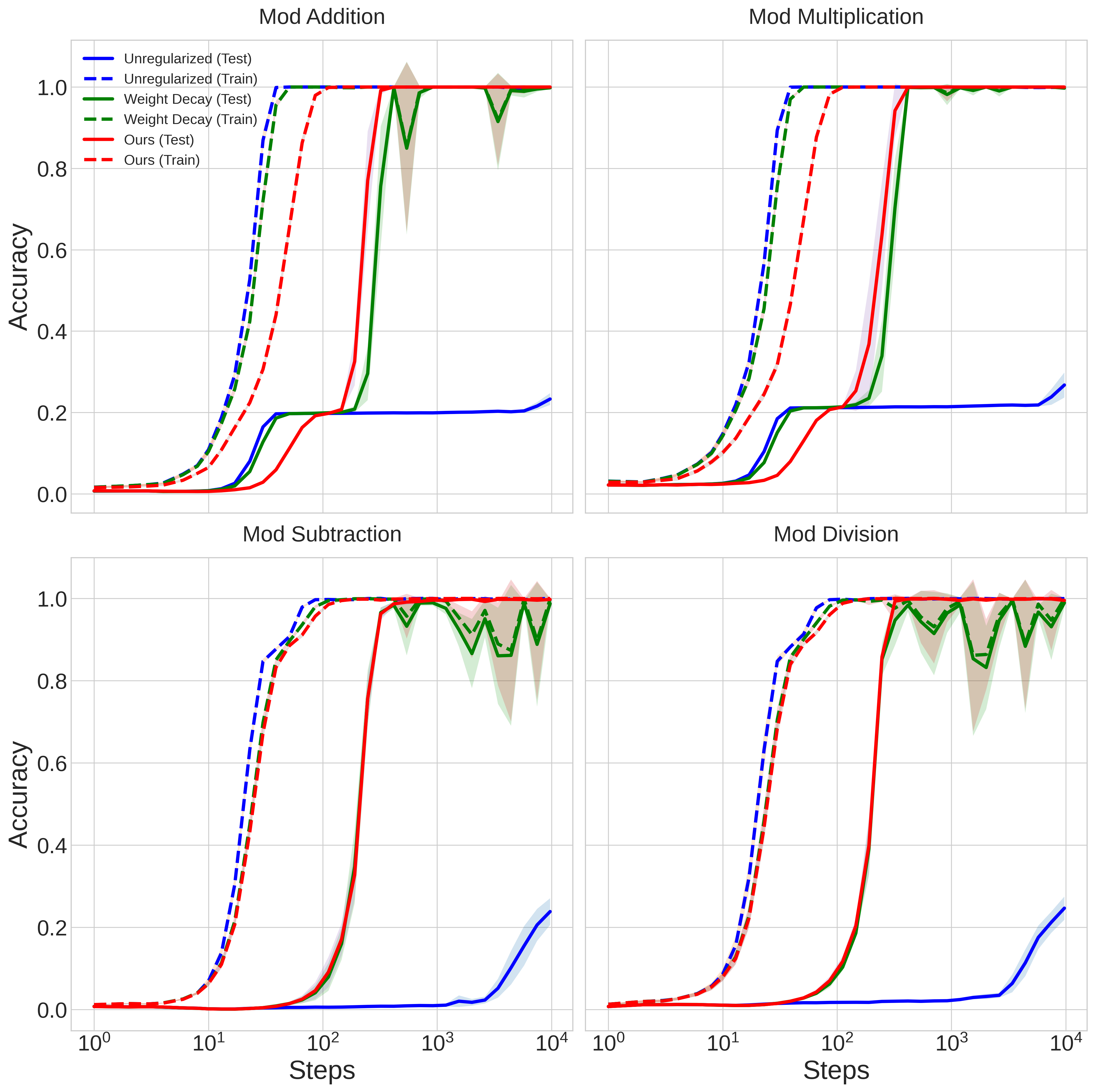}
    \caption{Test and train accuracy vs steps. Grokking occurs in all regularized models, and does not occur in unregularized models. We plot the mean over six seeds, and shade the std. error.}
    \label{accuracy}
\end{figure}

\begin{figure}
    \centering
    \includegraphics[width=0.9\linewidth]{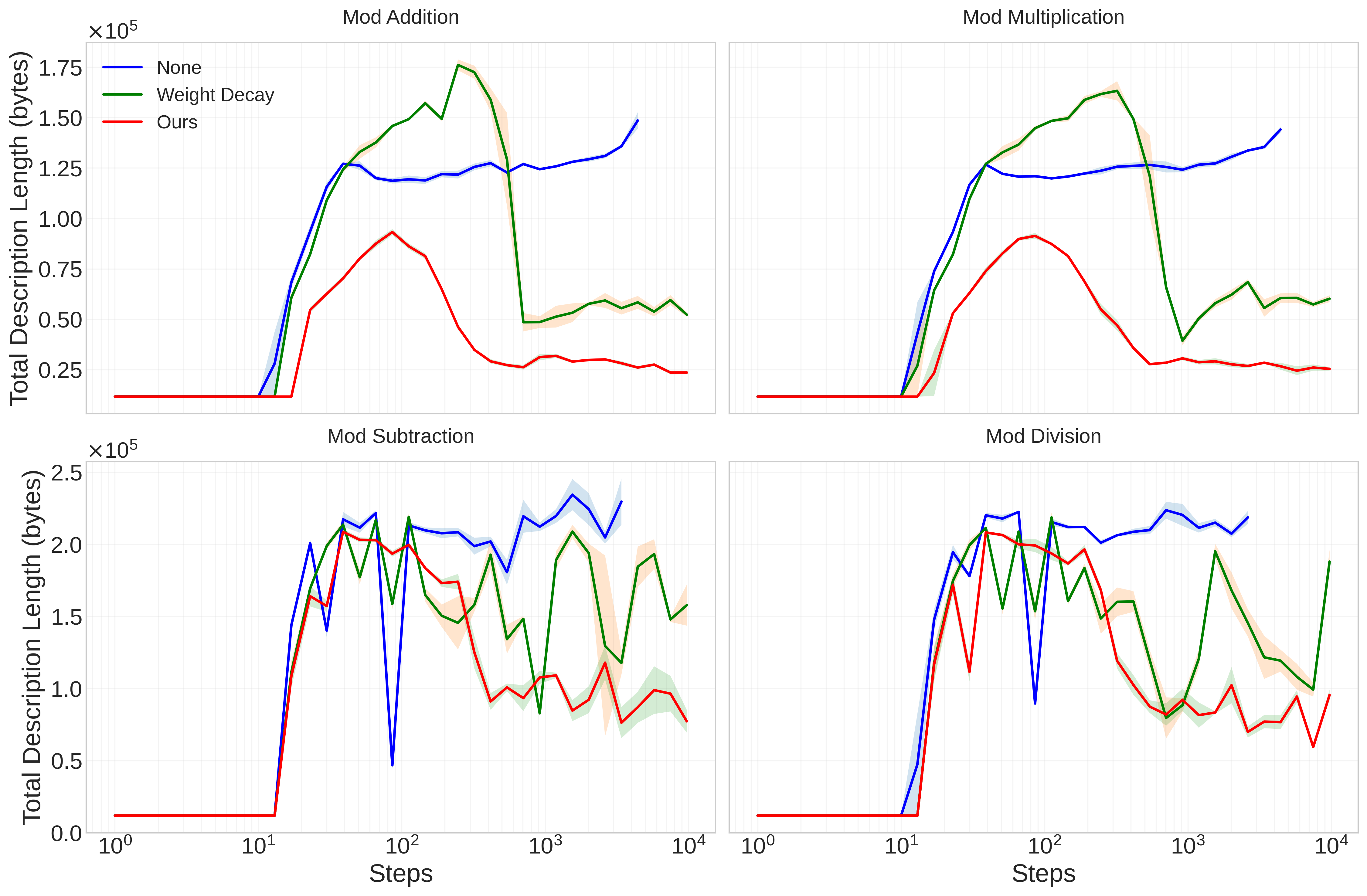}
    \caption{Total description length (complexity + entropy) vs training steps. Models trained with our regularization (red) have a smaller total description length compared to baselines. The unregularized model collapses during memorization in the subtraction and division tasks. Apparent complexity computed for $\epsilon=1$.}
    \label{TDL}
\end{figure}

\begin{table}[h]
\caption{Hyperparameters}
\label{hyperparams}
\begin{center}
\begin{tabular}{lll}
\multicolumn{1}{c}{\bf Description}  &\multicolumn{1}{c}{\bf Symbol} &\multicolumn{1}{c}{\bf Value}
\\ \hline \\
Modulus &$p$    &$113$\\
Dataset size         &$N$  &$p^2 = 12769$\\
Modular arithmetic tasks    &$\{+, \times, -,\div\}$    &-\\
Train data fraction per task         &-  &$(+, \times: 20\%),(-,\div: 30\%)$\\
Transformer layers  &$n_{\text{layers}}$   &$(+, \times: 1),(-,\div: 2)$\\
Transformer hidden dimension &$d_{\text{model}}$     &$128$\\
Transformer heads dimension     &$d_{\text{head}}$  &$32$\\ 
Transformer head count  &$n_{\text{head}}$  &$4$\\
Transformer MLP dimension   &$d_{\text{mlp}}$   &$512$\\
Noise regularization scale &$\delta$   &$(+, \times: 10^{-2}),(-,\div: 10^{-3})$\\
Spectral entropy scalar &$\beta$    &$10^{-1}$\\
Optimizer         &-  &AdamW\\
Weight decay scalar &-  &$1$\\
Learning rate         &-  &$10^{-3}$\\
\end{tabular}
\end{center}
\end{table}

\end{document}